\documentclass[preprint,12pt]{elsarticle}

\usepackage{lineno}

\usepackage{subcaption}
\usepackage{amsmath}
\usepackage{amssymb}
\usepackage{amsthm}
\usepackage{mathtools}
\usepackage{stmaryrd}

\usepackage{algorithm, algorithmicx}
\usepackage{algpseudocode}

\usepackage{graphicx}

\usepackage{url}
\usepackage{hyperref}

\usepackage{booktabs}
\usepackage{tabularx}
\usepackage{multirow}
\usepackage[table]{xcolor}
\usepackage{soul}

\definecolor{Gray}{gray}{0.9}
\newcolumntype{g}{>{\columncolor{Gray}}c}
\DeclarePairedDelimiter\ceil{\lceil}{\rceil}
\definecolor{myred}{rgb}{0.8, 0.0, 0.0}
\definecolor{myblue}{rgb}{0.0, 0.4, 0.8}

\journal{Engineering Applications of Artificial Intelligence}

\begin{document}

\begin{frontmatter}



\title{Interpretable Anomaly Detection with DIFFI: Depth-based Isolation Forest Feature Importance}


\author[1,2]{Mattia~Carletti}
\ead{mattia.carletti@unipd.it}
\author[2]{Matteo~Terzi}
\ead{matteo.terzi@unipd.it}
\author[1,2]{Gian~Antonio~Susto}
\ead{gianantonio.susto@unipd.it}

\address[1]{Human Inspired Technology Research Centre, University of Padua, Italy.}
\address[2]{Department of Information Engineering, University of Padua, Italy.}

\begin{abstract}
Anomaly Detection is an unsupervised learning task aimed at detecting anomalous behaviours with respect to historical data. In particular, multivariate Anomaly Detection has an important role in many applications thanks to the capability of summarizing the status of a complex system or observed phenomenon with a single indicator (typically called `Anomaly Score') and thanks to the unsupervised nature of the task that does not require human tagging. The Isolation Forest is one of the most commonly adopted algorithms in the field of Anomaly Detection, due to its proven effectiveness and low computational complexity. A major problem affecting Isolation Forest is represented by the lack of interpretability, an effect of the inherent randomness governing the splits performed by the Isolation Trees, the building blocks of the Isolation Forest. In this paper we propose effective, yet computationally inexpensive, methods to define feature importance scores at both global and local level for the Isolation Forest. Moreover, we define a procedure to perform unsupervised feature selection for Anomaly Detection problems based on our interpretability method; such procedure also serves the purpose of tackling the challenging task of feature importance evaluation in unsupervised anomaly detection. We assess the performance on several synthetic and real-world datasets, including comparisons against state-of-the-art interpretability techniques, and make the code publicly available to enhance reproducibility and foster research in the field.
\end{abstract}




\end{frontmatter}


\section{Introduction}
\label{sec:introduction}


Anomaly Detection (AD) techniques aim at automatically identifying anomalies (or outliers) within a given collection of data points \cite{borghesi2019semisupervised}. Their effectiveness is of paramount importance in a wide array of application domains, ranging from wireless sensor networks \cite{miao2018distributed} and industrial cyber-physical systems \cite{puggini2018enhanced, yang2017anomaly} to healthcare \cite{meneghetti2018data}, driving systems \cite{zhang2017safedrive} and biology \cite{zhang2020scale}. Their high usability is mainly due to the fact that most AD algorithms can be trained and deployed in unsupervised settings. This is particularly useful in environments where the data labelling process by human experts is prohibitively expensive and time-consuming, calling for an human-centered design principle that guarantees the minimization of human efforts. 

In recent years, a growing volume of research has been focusing on approaches based on Deep Neural Networks (DNNs) to tackle the AD task, especially for applications involving graphs \cite{yuan2015hyperspectral} and videos \cite{sabokrou2017deep}. Despite the high performance, DNNs cannot be considered as the ultimate solution to any AD problem as they exhibit a number of drawbacks in several real-world scenarios: i) depending on the complexity of the task and the dimensionality of the data, the training process of a DNN might last many hours or even days; ii) state-of-the-art DNN models are implemented (and trained) on expensive Graphics Processing Units, that might not be affordable in environments/applications characterized by limited budget or resource-constrained devices; iii) typically, a huge number of data points are required for the DNN to get satisfying generalization capabilities. For these reasons, there still persists a countless number of applications where traditional AD techniques, such as LOF \cite{breunig2000lof}, ABOD \cite{kriegel2008angle}, Isolation Forest \cite{liu2008isolation, liu2012isolation}, are being preferred over solutions based on DNNs.

Although AD algorithms have proved to be extremely useful and effective, their widespread adoption is far from being a reality even in industries and organizations with adequate infrastructures. This is actually a more general problem affecting any technology based on Machine Learning (ML) and it is mainly due to two `soft' factors:
(i) lack of confidence/trust from the users in AD algorithm outcomes and (ii) not immediate association between AD algorithm outcomes and root causes.
The first issue arises from the lack of labelled data points (that, on the other hand, is one of the main reasons why AD algorithms are appealing in the first place), which makes it impossible to set up an adequate testing procedure. This leads either to blindly trust the algorithm or not to use it at all, both the cases being undesirable. The second question, instead, investigates the possibility to gain additional knowledge about the task at hand, which may translates into actionable insights for troubleshooting or root cause analysis. The aforementioned issues can be addressed following the principles of eXplainable Artificial Intelligence (XAI) \cite{gunning2017explainable}, whose objective is to make so-called \textit{black-box} ML models easily understandable by human beings.

In the remainder of this Section we review the relevant literature in the XAI field (Section \ref{sec:literature}), provide an overview of the main contributions of this work (Section \ref{sec:contributions}) and the motivations at its core (Section \ref{sec:motivations}). Section \ref{sec:DIFFI} is devoted to the description and analysis of the proposed interpretability methods while in Section \ref{sec:results} the experimental results and a discussion thereof are provided. Finally, in Section \ref{sec:conclusion} we draw the conclusions and identify some interesting research directions for future works.

\subsection{Related works}
\label{sec:literature}
The general goal of XAI is to shed light on the inner workings of Machine Learning and Deep Learning models, especially in the context of regression and classification problems. A major focus is put on DNNs \cite{che2016interpretable} and ensemble methods \cite{tolomei2017interpretable}, two emblematic examples of algorithms classes that provide models that are highly accurate, but really hard to be understood by humans.

Given the fact that DNNs achieve state-of-the-art performance on several complex tasks such as image classification, text classification and time series forecasting - just to name a few - it comes as no surprise that a considerable volume of research in the XAI field focused on the problem of DNNs interpretability. The latter can be tackled with the purpose of either providing explanations about the predictions (i.e. the outputs) produced by the model \cite{zeiler2014visualizing, simonyan2013deep, murdoch2018beyond}, or interpreting the internal representations of the processed data \cite{sharif2014cnn, bau2017network}. It is worth highlighting a third promising line of research aimed at designing inherently interpretable DNNs \cite{melis2018towards, li2018deep, oreshkin2019n}. Since a complete dissertation on DNNs interpretability is out of the scope of this work, we refer the interested readers to \cite{gilpin2018explaining, zhang2018visual}.

As regards ensemble methods, we mainly relate our work to Random Forests (RFs) \cite{breiman2001random}, but several works on the interpretation of other ensembles (such as Gradient Boosting Decision Trees) can be found in literature \cite{valdes2016mediboost, wang2018tem}. RFs are ensembles of classification or regression trees leveraging \textit{bagging} to reduce the variance of predictions. With respect to single Decision Trees, RFs significantly improve performance in terms of accuracy, at the price of reduced interpretability. In this context, many works address the problem of improving standard feature importance score methods. Relevant examples are \cite{strobl2008conditional}, which proposes an improvement of the permutation importance measure based on a conditional permutation scheme, and \cite{li2019debiased}, in which the authors introduce a variant of the Mean Decrease Impurity (MDI) feature importance measure aimed at overcoming the problem of MDI feature selection bias. Besides single-feature importance measures, it is worth mentioning some recent works focused on the detection of interactions between features \cite{basu2018iterative, lundberg2018consistent, devlin2019disentangled}. 

The interpretability methods mentioned above are tailored for particular ML models and are dubbed \emph{model-specific}. They are characterized by high \emph{translucency} \cite{molnar2019interpretable}, i.e. they heavily rely on the inherent structure of the specific ML model under examination. There also exist several flexible approaches which earned remarkable interest due to their high \emph{portability} (i.e. they can be applied to a wide range of models), so-called \emph{model-agnostic} methods. Among the most prominent model-agnostic techniques used to explain individual predictions two popular approaches are LIME \cite{ribeiro2016should} and SHAP \cite{lundberg2017unified}. Instead, Partial Dependence Plots \cite{friedman2001greedy} and Accumulated Local Effects plots \cite{apley2016visualizing} represent examples of model-agnostic methods used to explain the model's behavior at a global level. While on one hand high portability may appear as an attractive feature for interpretability methods, on the other hand the interpretability problem is usually dealt with only once a specific model type has been chosen and the usefulness of model-agnostic methods simply lies in the lack of model-specific methods for several ML models classes. Moreover, model-agnostic methods usually exhibit not negligible shortcomings:
\begin{itemize}
    \item Since the inner structure of the model being examined is not exploited, the user might suspect that the provided explanation is just a simplistic and coarse approximation of the true underlying relation between the input and the output.
    \item The majority of model-agnostic methods are based on the manipulation of inputs and evaluation of the effects said manipulations induce on the corresponding predictions. This represents a delicate process as the artificially created input instances might not belong to the original data manifold, potentially causing stability issues and raising doubts about the actual information conveyed by the interpretability method. 
    \item In light of the need for further restrictive assumptions and/or opaque methodological choices (e.g. independence between features, the creation of perturbed input instances), the user is asked to take a leap of faith and consider the method as reasonable while not fully understanding the theoretical underpinnings. This simply shifts the problem from the lack of trust in the model to the lack of trust in the interpretability method itself.
\end{itemize}
Exhaustive descriptions, analyses and examples of both model-specific and model-agnostic approaches can be found in \cite{guidotti2019survey, molnar2019interpretable}.

\subsection{Contributions}
\label{sec:contributions}
Motivated by its ability to attract the attention of a growing and heterogeneous community of researchers and practitioners, we directed our efforts to the interpretation of the Isolation Forest (IF) \cite{liu2008isolation,liu2012isolation}. The IF model is particularly appreciated and widely used thanks to its high detection performance (often even with default hyperparameters values, with no tuning required) and its computational efficiency. Despite that, just like all ensemble learning methods, it might trigger perplexities and doubts as far as interpretability is concerned: indeed, no information about the logic behind the mechanism producing the predictions is available and neither an indication about which are the most relevant features to solve the AD task. In this work, we propose for the first time \emph{model-specific} methods (i.e. methods based on the particular structure of the IF model) to address the mentioned issues. Specifically, we introduce:
\begin{itemize}
    \item A global interpretability method, called \emph{Depth-based Isolation Forest Feature Importance (DIFFI)}, to provide Global Feature Importances (GFIs) which represent a condensed measure describing the macro-behavior of the IF model on \emph{training} data.
    \item A local version of the DIFFI method, called \emph{Local-DIFFI}, to provide Local Feature Importances (LFIs) aimed at interpreting individual predictions made by the IF model at \emph{test} time. 
    \item A simple and effective procedure to perform unsupervised feature selection for AD problems based on the DIFFI method.
\end{itemize}

Each contribution mentioned above complies with the human-centered principle adopted throughout this work, whose main goal is to match the user's needs to the best extent possible. This translates into a number of characteristics we sought to prioritise, e.g. limited computational times, light and straightforward hyperparameters tuning procedures. Additionally, our approach does not require additional knowledge (e.g. game theory concepts, necessary to fully grasp the rationale behind the functioning of SHAP), since it is based on very basic computations on quantities that naturally emerge from the principles governing the IF model. Along these lines, the proposed methods are consistent with the simplicity that characterizes the IF model, thus avoiding the risk of developing an interpretability framework which is more complex than the model itself.

\subsection{Motivations}
\label{sec:motivations}
If we consider the design of the evaluation procedure as part of the problem formalization process, the need for interpretable algorithms in the context of AD is consistent with the connection between the notions of interpretability and incompleteness evidenced in \cite{doshi2017towards}. Indeed, due to the lack of labelled datasets in AD problems, AD algorithms are practically rarely testable in unsupervised settings. To fill this gap that may prevent the adoption of such automated systems, we need to provide proxies to assess their trustworthiness. 

DIFFI is, to the best of our knowledge, the first model-specific method addressing the need for interpretability for the IF detector. The global DIFFI method is inspired by the preliminary work \cite{carletti2019explainable}, but differs entirely in the information is supposed to convey. While in \cite{carletti2019explainable} the goal is to get additional knowledge on the specific AD problem at hand (which is extremely useful especially in contexts where no domain expertise is available), in this work we focus on providing additional information about a trained instance of the IF model, with the main goal of increasing users' trust. Indeed, if the estimated feature importance scores aligned well with human prior knowledge, users would be more prone to lessen the supervision and safely give more autonomy to the machine (at least in non-critical scenarios). In this way, we could promote the adoption of the IF in fields where the professionals' skepticism towards intelligent algorithms is still a major obstacle to a massive use.

The model-specific nature of DIFFI is motivated by the will to reflect the actual logic governing the IF behavior and this may not be feasible with some model-agnostic techniques. For example, when exploiting interpretable surrogate models \cite{molnar2019interpretable} trained to approximate the predictions of a black box model, we need to make sure that the surrogate model fits the predictions of the original model with a satisfactory level of accuracy. Such a requirement represents an undesirable source of suspicion. Moreover, as argued in \cite{molnar2019interpretable, lipton2018mythos}, models commonly considered as universally interpretable, such as decision trees or linear regression, may lose their transparency advantage as they are asked to fit complex relations: very deep decision trees do not offer simple and intuitive visualizations, while linear regression is not suitable to model highly non-linear mappings.

DIFFI is a \textit{post-hoc method}: we decided to preserve the performance of an established and effective AD algorithm and focus on providing global and local feature importance measures computed a posteriori. The design of an intrinsically interpretable model would have required to sacrifice some predictive power in light of the trade-off between accuracy and interpretability \cite{molnar2019interpretable}.

The introduction of a local variant of the original algorithm for the interpretation of individual predictions serves a two-fold objective: i) it enables the interpretation of single data points in online settings, when the model has already been deployed, and ii) it helps in enhancing trust as the user can check not only whether the model tends to make mistakes on those kinds of inputs where humans also
make mistakes \cite{lipton2018mythos}, but also whether the misclassified inputs are being misinterpreted in the same way a human would.  

Finally, it should be noted that by providing both a global and a local interpretability method we can guarantee maximum flexibility: based on the required granularity or the amount of time that can be invested in the analysis of the results, the user has the possibility to choose the solution better suited to the specific scenario in which she operates.

\section{DIFFI: Depth-based Isolation Forest Feature Importance}
\label{sec:DIFFI}

In this Section, we first summarize the key concepts at the core of the IF algorithm and introduce the necessary notation. Then we extensively discuss the rationale behind the DIFFI method and thoroughly analyse each building block. We then propose a local variant of the DIFFI approach, \emph{Local-DIFFI}, for the interpretation of individual predictions. We conclude with the introduction of a novel method that is based on global DIFFI for unsupervised features selection; unsupervised feature selection can also be considered as a 'proxy' for evaluating the quality of attributed feature importance in unsupervised AD tasks.

\subsection{Background: the Isolation Forest}
\label{sec:background}


As introduced in Section \ref{sec:introduction}, the IF is an unsupervised AD algorithm leveraging an isolation procedure to infer a measure of outlierness, called \textit{anomaly score}, for each data point: the isolation procedure is based on recursive partitioning and aims at defining a region in the data domain where only the data point under examination lies. The underlying mechanism of IF is based on the reasonable hypothesis that the isolation procedure for outliers requires a limited number of iterations, while the isolation of inliers generally needs a larger number of recursive partitions; we will provide a formal description of the IF algorithm in the following.

The IF is an ensemble of \textit{Isolation Trees (ITs)} $\lbrace t_1, \dots, t_T  \rbrace$, i.e. base anomaly detectors characterized by a tree-like structure. ITs are data-induced random trees, in which each internal node $v$ is associated with a randomly chosen splitting feature (denoted $f(v)$) and a randomly chosen splitting threshold (denoted $\tau(v)$). Data points associated with node $v$ undergo a split test: points for which the value of $f(v)$ is less than $\tau(v)$ are sent to the left child of $v$, the others to the right child.

Given a dataset $\mathcal{D} = \lbrace \mathbf{x}_1, \dots,  \mathbf{x}_n \rbrace$ of $p$-dimensional data points, each IT $t$ is assigned a subset $\mathcal{D}_t \subset \mathcal{D}$ (usually called \textit{bootstrap sample}) sampled from the original set and carries out an isolation procedure based on the split tests associated to the internal nodes. Bootstrap samples have the same predetermined size, i.e.
$$|\mathcal{D}_t| = \psi \quad \text{for } t=1, \dots, T.$$
Data points in $\mathcal{D}_t$ (called \emph{in-bag samples}, from the perspective of tree $t$) are recursively partitioned until either all points are isolated or the IT reaches a predetermined depth limit $h_{max} = \ceil*{\log_2 (\psi)}$, function of the bootstrap samples size $\psi$. As a result, each data point $\mathbf{x}_i$ ends up in a leaf node, denoted $l_t(\mathbf{x}_i)$. We will denote with $h_t(\mathbf{x}_i)$ the number of edges that $\mathbf{x}_i$ passes through in its path from the root node to the corresponding leaf node, which is equivalent to the depth of the leaf node $l_t(\mathbf{x}_i)$.

The procedure described above is iterated over all ITs, each of which is assigned a different bootstrap sample. The anomaly score for a generic data point $\mathbf{x}_i$ is then computed as
\begin{equation} \label{eq:as}
    s(\mathbf{x}_i) = 2^{-\displaystyle \frac{\bar{h}(\mathbf{x}_i)}{c(\psi)}}
\end{equation}
where $c(\psi)$ is a normalization factor given by 
\begin{equation}
    c(\psi)=\begin{cases} 2H(\psi-1)-\frac{2(\psi-1)}{\psi} & \mbox{if }\psi>2,\\ 
    1 & \mbox{if }\psi=2,\\
    0, & \mbox{otherwise}
\end{cases}
\end{equation}
and $H(k)$ is the harmonic number which can be estimated as $H(k) \approx \text{ln}(k) + 0.5772156649$. $\bar{h}(\mathbf{x}_i)$ is the average path length associated with $\mathbf{x}_i$ and it is computed as 
\begin{equation}
    \bar{h}(\mathbf{x}_i) = \frac{1}{T} \sum_{t=1}^{T} h_t(\mathbf{x}_i).
\end{equation}

In the last step of the IF algorithm, anomalous data points are flagged through a thresholding operation on the anomaly scores. In this way, it is possible to partition the original set $\mathcal{D}$ as follows:
\begin{itemize}
    \item the subset of predicted inliers $\mathcal{P}_I = \lbrace \mathbf{x}_i \in \mathcal{D} \, | \, \hat{y}_i = 0  \rbrace$,
    \item the subset of predicted outliers $\mathcal{P}_O = \lbrace \mathbf{x}_i \in \mathcal{D} \, | \, \hat{y}_i = 1  \rbrace$,
\end{itemize}
where $\hat{y}_i \in \lbrace 0,1 \rbrace$ is the binary label produced by the thresholding operation, indicating whether the corresponding data point $\mathbf{x}_i$ is anomalous ($\hat{y}_i=1$) or not  ($\hat{y}_i=0$).

For further details on the IF algorithm and its properties, we refer the reader to the original paper \cite{liu2008isolation} and to the extended work \cite{liu2012isolation}. To conclude, it is worth highlighting that the IF, as a tree-based ensemble model, shares an inherent structure similar to that of RF. Nonetheless, random choices in IF have a far greater impact since, unlikely in RF, attributes associated with internal nodes are not selected according to specific splitting criteria but, indeed, randomly. This may be daunting to researchers interested in making the IF interpretable, but the present work serves as evidence that finding a solution to such a challenge is feasible. 

\subsection{DIFFI}
\label{sec:DIFFI_global}

The DIFFI method relies on two simple hypothesis exploited to define feature importance for the AD task at hand. A split test associated with a feature deemed as important should: 
\begin{itemize}
    \item[I1)] induce the isolation of anomalous data points at small depths (i.e. close to the root), while relegating regular data points to the bottom end of the trees;
    \item[I2)] produce higher imbalance on anomalous data points, while ideally being useless on regular points.  
\end{itemize}


Let us consider a trained instance of the IF detector and the corresponding training set $\mathcal{D} = \lbrace \mathbf{x}_1, \dots,  \mathbf{x}_n \rbrace$. 
For each tree $t$ we partition, based solely on the predictions produced by tree $t$, the assigned bootstrap sample $\mathcal{D}_t$ into the subset of predicted inliers $\mathcal{P}_{I,t}$ and the subset of predicted outliers $\mathcal{P}_{O,t}$, where 
\begin{align*}
    \mathcal{P}_{I,t} &= \lbrace \mathbf{x}_i \in \mathcal{D}_t \, | \, \hat{y}_{i,t} = 0  \rbrace, \\
    \mathcal{P}_{O,t} &= \lbrace \mathbf{x}_i \in \mathcal{D}_t \, | \, \hat{y}_{i,t} = 1  \rbrace,
\end{align*}
and $\hat{y}_{i,t}$ denotes the prediction produced by tree $t$ associated to $\mathbf{x}_i$. Predictions are obtained, as usual, through a thresholding operation on the anomaly scores, which are now computed by replacing $\bar{h}(\mathbf{x}_i)$ with $h_t(\mathbf{x}_i)$ in \eqref{eq:as}. The choice to consider only bootstrap samples for each tree, rather than the entire training set, is motivated by the need to reduce the computational cost and the desire of decoupling the evaluation of feature importance scores from the generalization capability of the trained model. Indeed, each tree is trained only on the corresponding bootstrap sample and by considering the whole training set the performance of the tree on unseen data (other than the bootstrap sample) might affect the resulting feature importance scores. In general, this is not a problem since training data are supposed to be drawn from the same distribution and, thus, the performance on in-bag and out-of-bag samples should be similar. Nonetheless, by considering only in-bag samples for each tree the computational cost can be made independent from the training set size. 

\begin{figure}[htb]
\centering
    \includegraphics[scale=0.38]{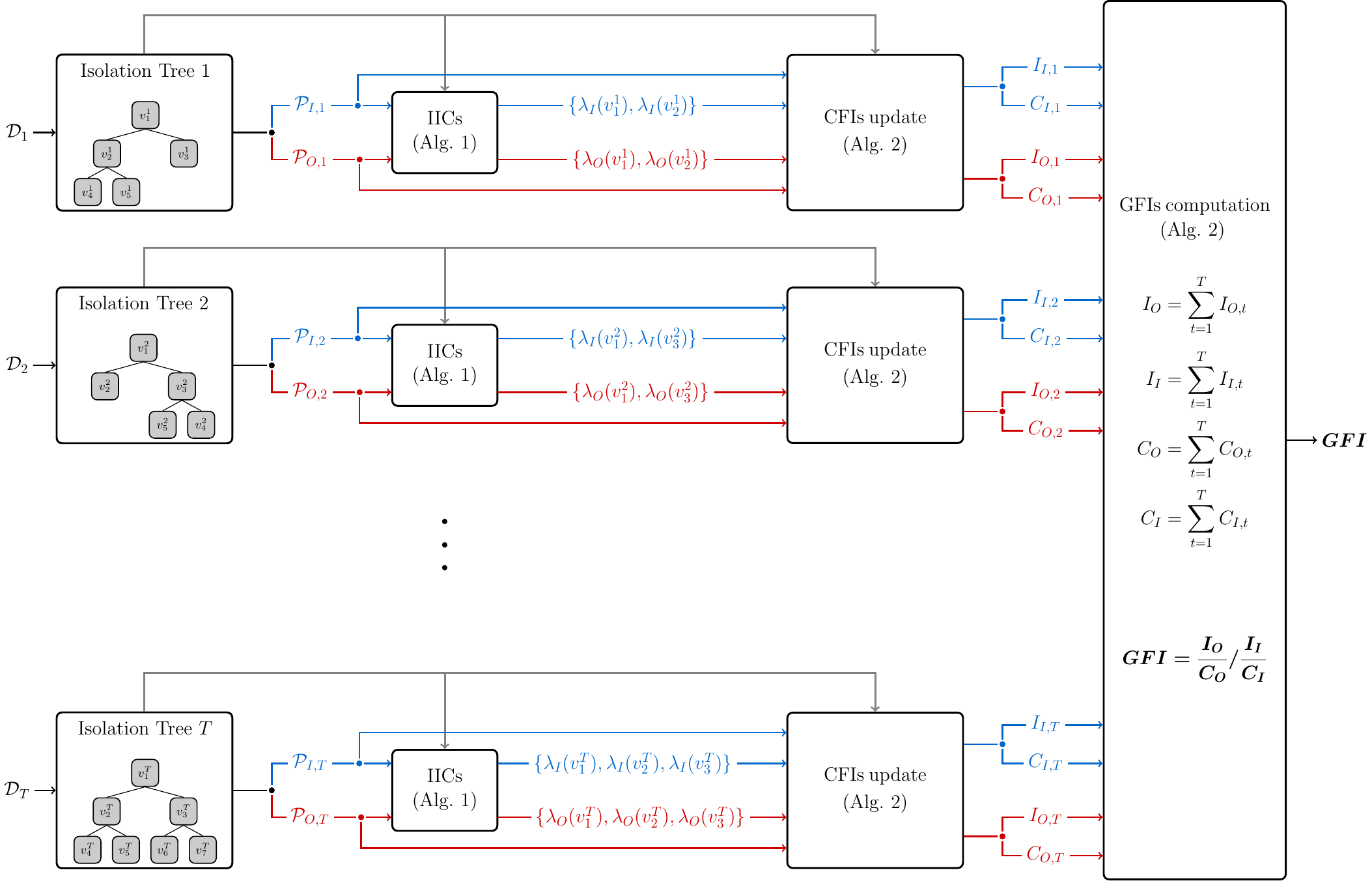}
    \caption{Overview of the DIFFI method for global feature importance scores. The notation $v_i^t$ denotes the $i$-th node in the $t$-th tree. The quantities and the information flow related to predicted \textcolor{myblue}{inliers}/\textcolor{myred}{outliers} are in \textcolor{myblue}{blue}/\textcolor{myred}{red}, respectively. Boxes denote computation blocks: more details on IICs and CFIs/GFIs computation will be provided in Alg. \ref{alg: DIFFI_IICs} and Alg. \ref{alg: DIFFI_GFIs} respectively. 
} 
    \label{fig:diffi_overview}
\end{figure}

We will define \textit{Cumulative Feature Importances (CFIs)} for inliers and outliers, real-valued quantities that will be then properly normalized and combined together to produce the final feature importance measures. CFIs are updated in an additive fashion by exploiting data points in $\mathcal{P}_{I,t}$ and $\mathcal{P}_{O,t}$, for $t=1, \dots, T$. The update rule depends on two quantities that reflect the two intuitions explained above: the depth of the leaf node where a specific data point ends up (intuition I1) and the \textit{Induced Imbalance Coefficient (IIC)} associated with a specific internal node (intuition I2). In the remainder of this Section we first explain how to compute IICs, then we describe the procedure for the update of CFIs for inliers and outliers and combine them to produce the final GFIs. An overview of the method and its building blocks is given in Figure \ref{fig:diffi_overview}.

\subsubsection{IICs computation}

Let us consider the generic internal node $v$ in tree $t$. Let $n(v)$ represent the number of data points associated with node $v$, $n_{l}(v)$ the number of data points associated with its left child and $n_{r}(v)$ the number of data points associated with its right child. The IIC of node $v$, denoted $\lambda(v)$, is obtained as follows
\begin{equation} \label{eq: iic}
\lambda(v)= \begin{cases} 0, & \mbox{if } n_l(v)=0 \mbox{ or } n_r(v)=0 \\ 
\\  \tilde{\lambda}(v), & \mbox{otherwise} \end{cases}
\end{equation} 
where
\begin{equation} \label{eq: iic_g}
\tilde{\lambda}(v) = g \left( \displaystyle\frac{\max (n_l(v), n_r(v))}{n(v)} \right)
\end{equation}
and $g(\cdot)$ is a scaling function mapping its input into the interval $\left[0.5, 1\right]$, whose rationale is detailed below. In \eqref{eq: iic}, the first case represents a \emph{useless split}, in which all data points are sent either to the left or right child. The best possible split, instead, is what we call an \emph{isolating split}: this happens when either the left or the right child receives exactly one data point. An isolating split is assigned the highest possible IIC, i.e. 1.

For the experiments we use the following scaling function
\begin{equation} \label{eq:norm}
    g(a) = \frac{a - \lambda_{min}(n)}{\lambda_{max}(n) - \lambda_{min}(n)} \cdot 0.5 + 0.5,
\end{equation}
where
$\lambda_{min}(n)$ and $\lambda_{max}(n)$ denote the minimum and maximum scores, respectively, that can be obtained a priori\footnote{Here ``a priori'' means before applying the scaling function $g(\cdot)$.} given the number $n(v)$ of data points associated to the specific node location $v$. We notice that by scaling the values we can mitigate undesired effects due to the exact value of $n(v)$, which should not affect the computation of the induced imbalance. For instance, if $n=10$ data points are associated to a specific node, the worst non-useless split ($5$ points to each child) leads to $\lambda_{min}=0.5$. Instead, if $n=11$, the worst non-useless split ($5$ samples to one child and $6$ to the other one) leads to $\lambda_{min}=\frac{6}{11}=0.54$. With the introduction of the scaling function $g(\cdot)$, the two scenarios depicted above are equivalent at least for the extreme cases of isolating split and worst non-useless split. 

As it will be clear later on, we need to distinguish between IICs for inliers, denoted $\lambda_I(v)$ and computed by exploiting the predicted inliers $\mathcal{P}_{I,t}$, and the counterpart for outliers, denoted $\lambda_O(v)$ and computed by exploiting the predicted outliers $\mathcal{P}_{O,t}$. 

The computation of IICs is performed for each internal node $v$ of each tree $t = 1, \ldots, T$ in the forest. The pseudocode is reported in Algorithm \ref{alg: DIFFI_IICs}.

\alglanguage{pseudocode}
\begin{algorithm}[tb]
	\caption{DIFFI - IICs computation.}
	\hspace*{\algorithmicindent} \textbf{Input:} Isolation Tree $t$, set of data points $\mathcal{S}$\\
    \hspace*{\algorithmicindent} \textbf{Output:} IICs vector $\Lambda \in \mathbb{R}^{|t|}$ ($|t|$ number of internal nodes in tree $t$)
	\begin{algorithmic}[1]
		\vspace{1mm}
		\Function{IIC}{$t, \mathcal{S}$}
		\For{internal node $v$ in $t$}
		\State Compute $\lambda(v)$ as in Eq. \eqref{eq: iic}, using samples in $\mathcal{S}$
		\State $\Lambda[v] \gets \lambda(v)$
		\EndFor 	
		\State \Return $\Lambda$	
		\EndFunction
	\end{algorithmic}
	\label{alg: DIFFI_IICs}
\end{algorithm}

\subsubsection{CFIs update}

We distinguish between CFI for inliers, denoted $I_I$, and the counterpart for outliers, denoted $I_O$. Both $I_I$ and $I_O$ are $p$-dimensional vectors, where the $j$-th component represents the CFI (for inliers or outliers) for the $j$-th feature. 

Let $Path(\mathbf{x},t)$ be the path from the root node to the corresponding leaf node associated with data point $\mathbf{x}$ in tree $t$. For simplicity we restrict the attention to the CFI update rule for inliers (i.e. $I_I$), as the extension to $I_O$ is immediate.  First we initialize $I_I = \mathbf{0}_p$, where $\mathbf{0}_p$ denotes the $p$-dimensional vector of zeros. Then we update $I_I$ in an additive fashion. Specifically, we iterate over all trees in the forest and then, for each tree $t$, over the subset of predicted inliers $\mathcal{P}_{I,t}$. Finally, for the generic predicted inlier $\mathbf{x}_I \in \mathcal{P}_{I,t}$, we iterate over the internal nodes in its path $Path(\mathbf{x}_I,t)$. If the splitting feature associated with the generic internal node $v$ is $f_j$, then we update the $j$-th component of $I_I$ by adding the quantity
\begin{equation} \label{eq:cumul_impo_update}
	\Delta = \displaystyle \frac{1}{h_t(\mathbf{x}_I)} \cdot \lambda_I(v),
\end{equation}
where we recall that $h_t(\mathbf{x}_I)$ denotes the depth of the leaf node (in tree $t$) associated with data point $\mathbf{x}_I$.  In \eqref{eq:cumul_impo_update}, we can notice the contributions of two factors, formalizing the two intuitions at the core of DIFFI. Indeed, the right-hand side factor characterizes the ``local" effect of the split through the induced imbalance at that specific node location. The left-hand side factor, instead, characterizes the ``global" effect of the split taking into account potential situations in which an apparently bad (from the ``local" perspective) split actually re-organizes the data points in a way that makes it easier for subsequent split tests to isolate them.  

As regards the update rule for $I_O$, the only differences with respect to the procedure detailed above are that we iterate over $\mathcal{P}_{O,t}$ rather than $\mathcal{P}_{I,t}$ and that we replace $\lambda_I(v)$ with $\lambda_O(v)$ in \eqref{eq:cumul_impo_update}.

\subsubsection{GFIs computation}
Recalling that in the IF, differently to what happens in the RF algorithm, the splitting features are selected randomly, the careful reader should perceive a potential issue: if the generic feature $f$ was sampled more frequently than others, it would unfairly receive a higher CFI. 
We define the \emph{features counter} for inliers, denoted $C_I$, and the counterpart for outliers, denoted $C_O$, as $p$-dimensional vectors where the $j$-th component represents how many times the $j$-th feature appeared while updating the CFIs. $C_I$ is updated iterating over $\mathcal{P}_{I,t}$, while $C_O$ is updated iterating over $\mathcal{P}_{O,t}$, for $t=1, \ldots, T$.  
In order to filter out the effect of random splitting features selection, we simply normalize the CFIs by their corresponding features counters, $C_I$ and $C_O$ respectively. The GFIs are then obtained as
\begin{equation}\label{eq:fi}
    GFI = \frac{I_O/C_O}{I_I/C_I},
\end{equation}
where divisions are performed element-wise. Notice that higher feature importance for inliers (i.e. high value of the denominator in \eqref{eq:fi}) implies lower overall feature importance. This is consistent with intuition I1: important features isolate outliers closer to the root and simultaneously do not contribute to the isolation of inliers. Algorithm \ref{alg: DIFFI_GFIs} summarizes the whole procedure at the core of the DIFFI method. IICs are computed according to the function described in Algorithm \ref{alg: DIFFI_IICs}. The function $\text{Feat}(\cdot)$ returns the feature associated with the internal node passed as argument.

\alglanguage{pseudocode}
\begin{algorithm}[htb!]
	\caption{DIFFI - CFIs update and GFIs computation.}
	\hspace*{\algorithmicindent} \textbf{Input:} Isolation Forest F, bootstrap samples $\lbrace \mathcal{D}_1, \ldots, \mathcal{D}_T \rbrace$\\
    \hspace*{\algorithmicindent} \textbf{Output:} global feature importance scores $GFI \in \mathbb{R}^p$
	\begin{algorithmic}[1]
		\vspace{1mm}
		\State Initialize counter and cumulative importance variables:
		\For{$j=1, \dots, p$}
		\State $C_I(f_j), I_I(f_j), C_O(f_j), I_O(f_j) \gets 0$
		\EndFor
		\For{Isolation Tree $t$ in Isolation Forest F}
		\State Get predicted inliers and outliers according to tree $t$:
		\State $\mathcal{P}_{I,t}, \mathcal{P}_{O,t} \gets \text{Predict}(\mathcal{D}_t, t)$
		\State Get IICs associated with inliers and outliers:
		\State $\Lambda_I \gets \Call{IIC}{t, \mathcal{P}_{I,t}}$
		\State $\Lambda_O \gets \Call{IIC}{t, \mathcal{P}_{O,t}}$
		\State Update CFIs:
		\For{$\mathbf{x} \in \mathcal{P}_{I,t}$}
		\For{internal node $v$ in $\text{Path}(\mathbf{x}, t)$}
		\State $f=\text{Feat}(v)$
		\State $C_I(f) \gets C_I(f) + 1$
		\State $I_I(f) \gets I_I(f) + \displaystyle \frac{1}{h_t(\mathbf{x})} \cdot \Lambda_I[v]$
		\EndFor 		
		\EndFor
		\For{$\mathbf{x} \in \mathcal{P}_{O,t}$}
		\For{internal node $v$ in $\text{Path}(\mathbf{x}, t)$}
		\State $f=\text{Feat}(v)$
		\State $C_O(f) \gets C_O(f) + 1$
		\State $I_O(f) \gets I_O(f) + \displaystyle \frac{1}{h_t(\mathbf{x})} \cdot \Lambda_O[v]$
		\EndFor 		
		\EndFor
		\EndFor
		\State Compute FI for each feature:
		\For{$j=1, \dots, p$}
		\State $\text{GFI}(f_j) = \displaystyle \frac{I_O(f_j)}{C_O(f_j)} / \displaystyle \frac{I_I(f_j)}{C_I(f_j)}$
		\EndFor
	\end{algorithmic}
	\label{alg: DIFFI_GFIs}
\end{algorithm}

\subsection{Local-DIFFI}
\label{sec:DIFFI_local}

\alglanguage{pseudocode}
\begin{algorithm}[htb!]
	\caption{Local-DIFFI.}
	\hspace*{\algorithmicindent} \textbf{Input:} Isolation Forest F, predicted outlier $\mathbf{x}_O$\\
    \hspace*{\algorithmicindent} \textbf{Output:} local feature importance scores $LFI \in \mathbb{R}^p$
	\begin{algorithmic}[1]
		\vspace{1mm}
		\State Initialize counter and cumulative importance variables:
		\For{$j=1, \dots, p$}
		\State $C^{loc}_O(f_j), I^{loc}_O(f_j) \gets 0$
		\EndFor
		\For{Isolation Tree $t$ in Isolation Forest F}
		\State Update CFIs:
		\For{internal node $v$ in $\text{Path}(\mathbf{x}_O, t)$}
		\State $f=\text{Feat}(v)$
		\State $C^{loc}_O(f) \gets C^{loc}_O(f) + 1$
		\State $I^{loc}_O(f) \gets I^{loc}_O(f) + \displaystyle \frac{1}{h_t(\mathbf{x}_O)} - \frac{1}{h_{max}}$
		\EndFor
		\EndFor
		\State Compute FI for each feature:
		\For{$j=1, \dots, p$}
		\State $\text{LFI}(f_j) = \displaystyle \frac{I^{loc}_O(f_j)}{C^{loc}_O(f_j)}$
		\EndFor
	\end{algorithmic}
	\label{alg: DIFFI_local}
\end{algorithm}

For the interpretation of individual predictions produced by the IF, we exploit a procedure similar to the one described in Section \ref{sec:DIFFI_global} with differences due to the impossibility to compute some quantities in the local case (i.e. when considering one sample at a time). Specifically:
\begin{itemize}
    \item the Induced Imbalance Coefficients cannot be computed since we consider only one sample;
    \item all quantities referred to predicted inliers cannot be computed, since the focus is on the interpretation of predicted outliers.
\end{itemize}
Given a predicted outlier $\mathbf{x}_O$, the corresponding \emph{Local Feature Importance (LFI)} is computed as 
\begin{equation} \label{eq:fi_out}
    LFI(\mathbf{x}_O) = \frac{I^{loc}_O}{C^{loc}_O},
\end{equation}
where, similarly to the global case, $C^{loc}_O$ is the features counter and $I^{loc}_O$ is the CFI associated with $\mathbf{x}_O$. Both $C^{loc}_O$ and $I^{loc}_O$ are updated in an additive fashion by iterating over the ITs in the forest and over the internal nodes in the path $Path(\mathbf{x}_O,t)$ in each IT. Differently from the global case, if the splitting feature associated with the generic internal node $v$ (in tree $t$) is $f_j$, then we update the $j$-th component of $I^{loc}_O$ by adding the quantity
\begin{equation} \label{eq:delta_local}
    \Delta_{loc} = \frac{1}{h_t(\mathbf{x}_O)} - \frac{1}{h_{max}}.
\end{equation}
The correction term $- \displaystyle \frac{1}{h_{max}}$ in Equation \eqref{eq:delta_local} makes sure that CFI is not updated when the leaf node associated with the predicted outlier $\mathbf{x}_O$ is at the maximum depth of the tree. If the correction term was not used, the CFI of each feature in $Path(\mathbf{x}_O,t)$ would be updated also in cases where the data point under examination is not isolated (i.e. when it ends up in leaf nodes at the maximum depth) as $\Delta_{loc}$ would always be greater than zero.
The pseudocode of the Local-DIFFI method is reported in Algorithm \ref{alg: DIFFI_local}. The function $\text{Feat}(\cdot)$ returns the feature associated with the internal node passed as argument.

\subsection{Unsupervised feature selection with global DIFFI}
\label{sec:DIFFI_fs}

\begin{figure}[htb]
\centering
    \includegraphics[width=0.65\textwidth]{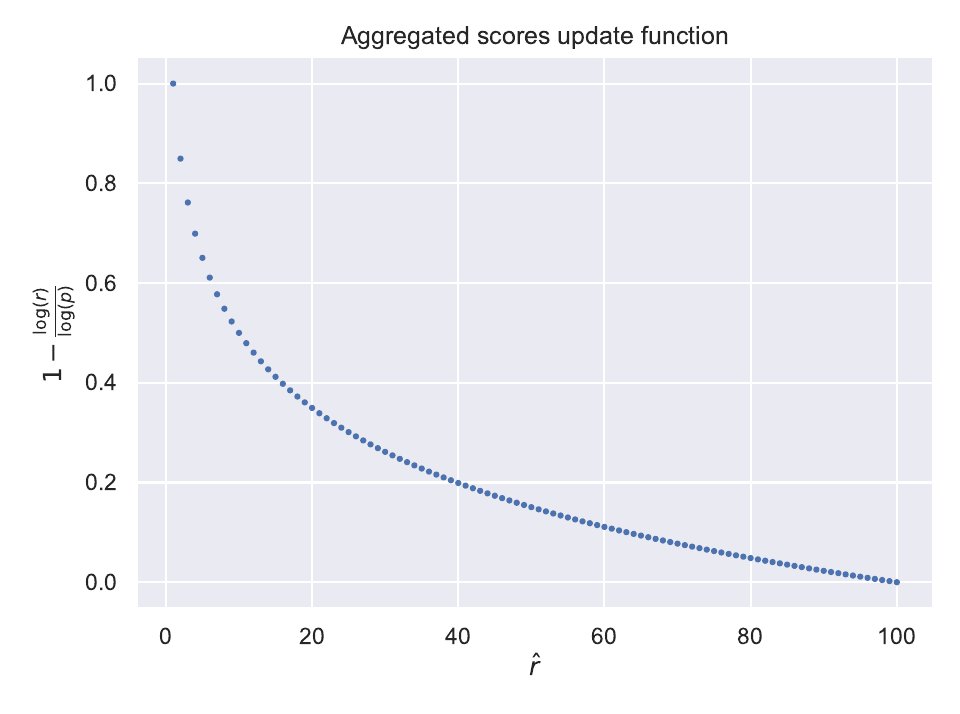}
    \caption{Update function for aggregated scores: example with $p=100$.}
    \label{fig:scores}
\end{figure}

The DIFFI method outlined in Section \ref{sec:DIFFI_global} can be effectively exploited to perform feature selection in the context of AD problems when labels associated to training data points are not available. The procedure consists in training $N_{fs}$ different instances of IF obtained with the same training data but different random seeds, in order to filter out effects due to the stochasticity inherently present in the model. The global DIFFI scores associated to each instance of IF are then aggregated to define a ranking on the features as follows:
\begin{enumerate}
    \item We define the $p$-dimensional vector of aggregated scores $S_{agg} \in \mathbb{R}^p$ initialized as a $p$-dimensional vector of zeros, where $p$ is the number of features.
    \item For each of the $N_{fs}$ IF instances: 
    \begin{itemize}
        \item we rearrange the global DIFFI scores in decreasing order, thus obtaining a ranking of the features (for the specific IF instance) from the most important one to the least important one;
        \item we update $S_{agg}$ by adding, for each feature, a quantity that is a function of the estimated rank $\hat{r}$, namely 
        \begin{equation} \label{eq:fs_agg}
            \Delta_{fs} = 1 - \frac{\log(\hat{r})}{\log(p)}.
        \end{equation}
        Notice that in \eqref{eq:fs_agg} we differentiate more the scores among most important features, while for the least important ones the scores are similar and very small (see Figure \ref{fig:scores}).
    \end{itemize} 
    \item The resulting vector of aggregated scores $S_{agg}$ is then used to define a ranking over the features: the higher the aggregated score the more important the feature.
\end{enumerate}

The motivations behind this strategy stem from the human-centered design principle adopted in this work and can be summarized as follows:
\begin{itemize}
    \item Users can spend only a limited amount of time on preprocessing operations since, especially in productive environments, the deployment of novel algorithmic solutions is usually meant to promptly react to emerging issues; the light computational cost of the DIFFI method, thanks to the in-bag samples trick, is particularly appealing in such time-constrained applications.
    \item With the intention of minimizing the effort of users, the choice of the IF (combined with DIFFI) as a proxy model to produce a ranking on the features is attractive as it introduces just a few hyperparameters to be tuned; in addition, it is worth mentioning that the IF is often preferred over other AD algorithms due to its good performance with the default hyperparameters values suggested in the original paper.
    \item The proposed strategy for unsupervised feature selection takes into account the nature of the task, while most of other methods do not. This is particularly important for AD tasks as relevant features for classification might not be relevant for AD. The user interested in solving an AD problem may trust more a method specifically suited for such purpose than other task-agnostic methods.
\end{itemize}

In addition to the unquestionable usefulness of the unsupervised feature selection task, the procedure outlined above also represents an excellent proxy to indirectly assess the quality of the feature importance scores provided by the global DIFFI method described in Section \ref{sec:DIFFI_global}. Indeed, good feature importance scores leads to a good ranking of the features, which in turn can be considered as a good solution to the unsupervised feature selection problem.

\section{Experimental Results}
\label{sec:results}

We report in this Section experimental results on synthetic and real-world datasets to assess the effectiveness of both the global DIFFI method to perform unsupervised feature selection and its local variant to provide feature importance scores associated to individual predictions.

We make the code publicly available\footnote{\url{https://github.com/mattiacarletti/DIFFI}} to enhance reproducibility of our experimental results and to foster research in the field.

\subsection{Interpretation of individual predictions}

For the interpretation of individual predictions provided by the IF model we exploit the local variant of the DIFFI method described in Section \ref{sec:DIFFI_local}. We assess the effectiveness of the Local-DIFFI method on a synthetic dataset and a real-world dataset: on both datasets we have prior knowledge about the most relevant features for the AD task to be solved, which would be fundamental for evaluating the performance of DIFFI. We remark how finding real-world data for AD tasks with a priori knowledge on the relevant features is not a trivial task. The experimental setup adopted here simulates a real scenario of remarkable interest in several application domains: given a trained instance of the IF, the user is interested in deploying the model in online settings to get the prediction and the corresponding local feature importance scores associated with the individual data point being processed. Typical applications include (but are not limited to) the monitoring of smart manufacturing systems and the detection of abnormal patterns in healthcare data: in both examples the promptness of responses may be crucial to ensure quick and effective corrective actions for the industrial processes/machines or for the well-being of patients.

\subsubsection{Synthetic dataset}

The synthetic dataset employed in this work was created by initially considering 2-dimensional data points whose dimension is then augmented by adding noise features, similarly to what done in \cite{carletti2019explainable}. Specifically, the generic data point $\mathbf{x}_i$ is represented by the $p$-dimensional vector
\begin{equation}
    \mathbf{x}_i = \left[ \rho \cos (\theta), \rho \sin (\theta), n_1, \dots, n_{p-2} \right]^\intercal,
\end{equation}
where 
$n_j \sim \mathcal{N}(0, 1)$  for $j=1, \dots, p-2$ are white noise samples.
Parameters $\rho$ and $\theta$ are random variables drawn from continuous uniform distributions. In particular, for regular data points we have
\begin{equation}
    \theta \sim \mathcal{U}(0, 2 \pi), \qquad \rho \sim \mathcal{U}(0, 3),    
\end{equation}
while for anomalous data points we have
\begin{equation}
    \theta \sim \mathcal{U}(0, 2 \pi), \qquad \rho \sim \mathcal{U}(4, 30).    
\end{equation}
For our experiments we consider a training set composed of 1000 $6$-dimensional data points (thus 4 noise features), with 10 \% anomalies. We trained an instance of IF with 100 trees and $\psi = 256$ (that are typical choices for the IF hyperparameters \cite{liu2008isolation}), and we obtained an F1-score on training data equal to 0.76.

\begin{figure}[tb]
\centering
    \includegraphics[width=0.65\textwidth]{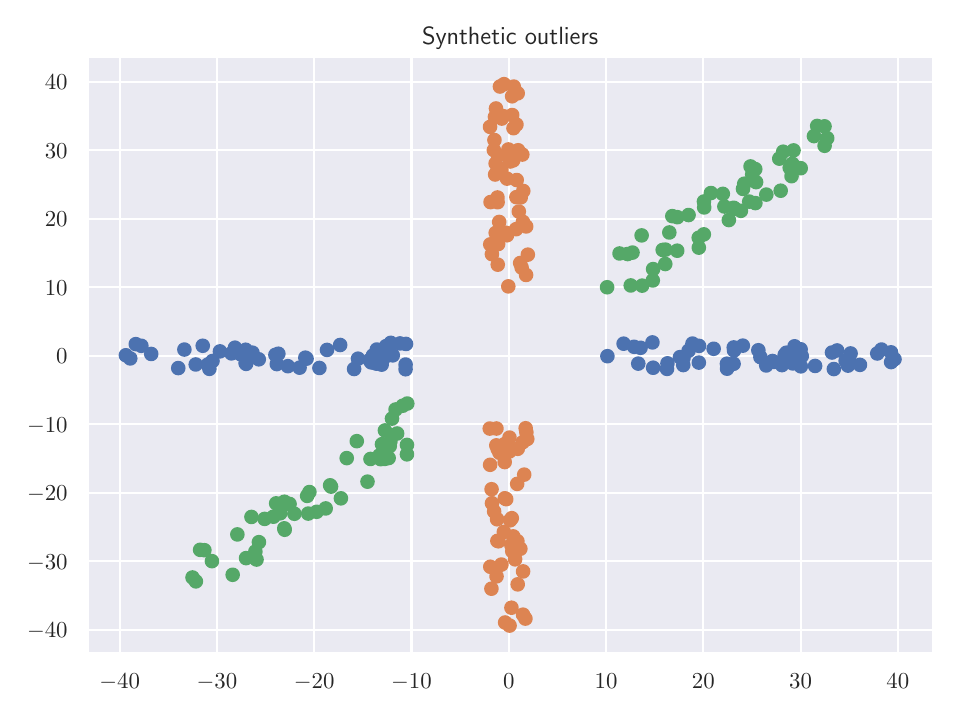}
    \caption{Synthetic outliers projected on the $f_1-f_2$ plane.}
    \label{fig:outliers}
\end{figure}

For the testing phase, we generated 300 additional ad-hoc anomalies, displayed in Figure \ref{fig:outliers} (projected on the subspace of relevant features): 100 lying on the $x$-axis (blue points), 100 on the $y$-axis (orange points) and 100 on the bisector (green points). The prior knowledge for this AD task is represented by the fact that only feature $f_1$ is relevant for outliers on the $x$-axis, only feature $f_2$ is relevant for outliers on the $y$-axis and both $f_1$ and $f_2$ are relevant for outliers on the bisector (all the other features, being white noise samples, are irrelevant in all cases).

\begin{figure}[]
\centering
\captionsetup[subfigure]{justification=centering}
	\begin{subfigure}[]{0.5\linewidth}
		\includegraphics[width=\textwidth]{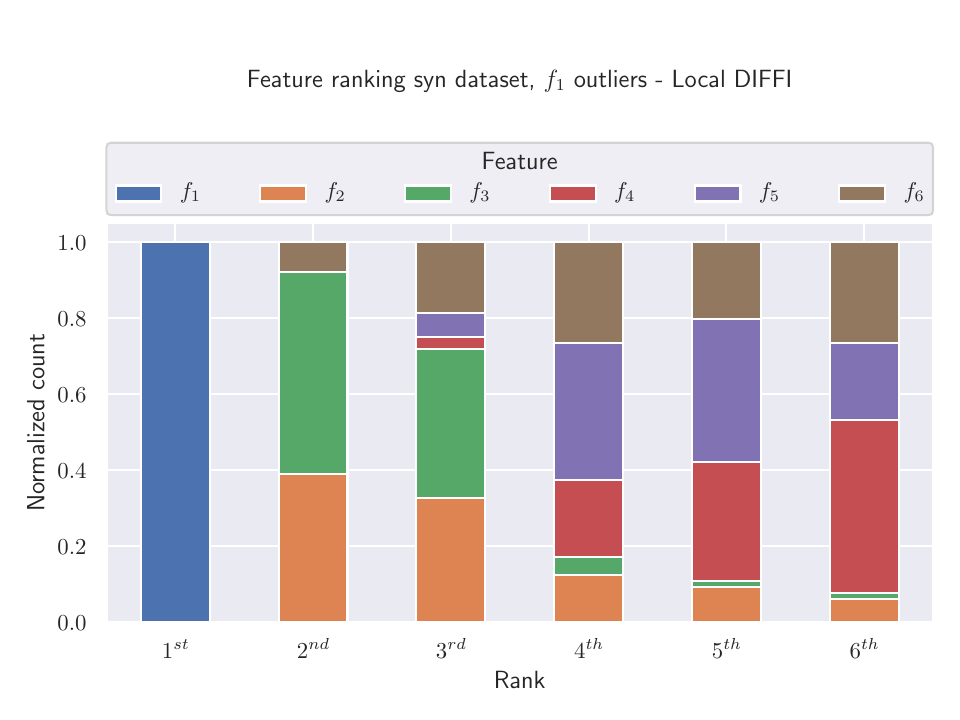}
	\end{subfigure}\hfill
	\begin{subfigure}[]{0.5\linewidth}
		\includegraphics[width=\textwidth]{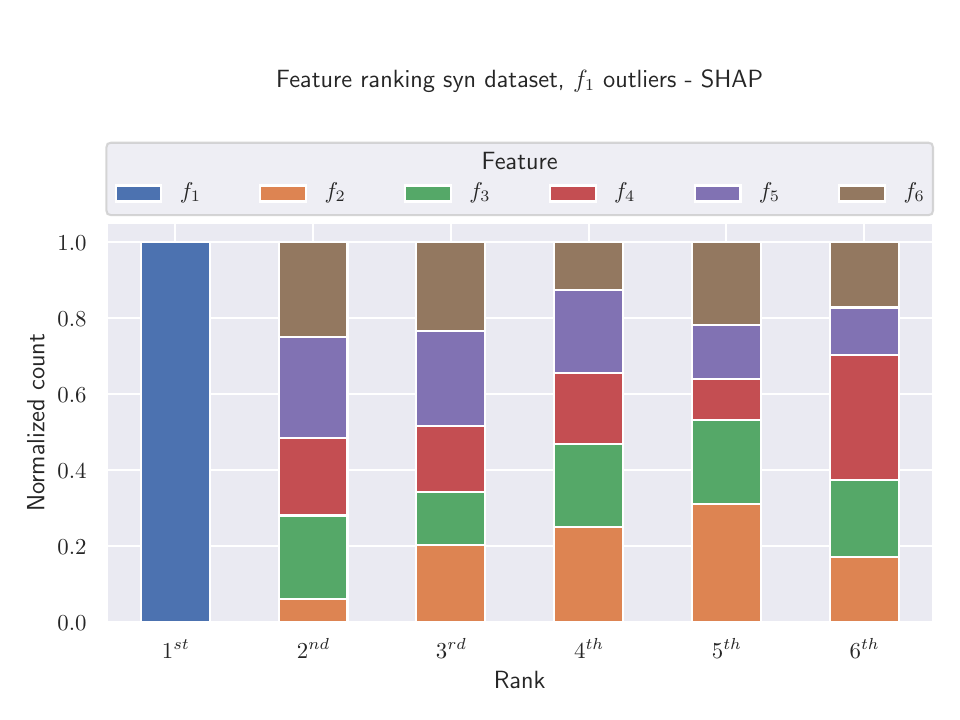}
	\end{subfigure}
	\begin{subfigure}[]{0.5\linewidth}
		\includegraphics[width=\textwidth]{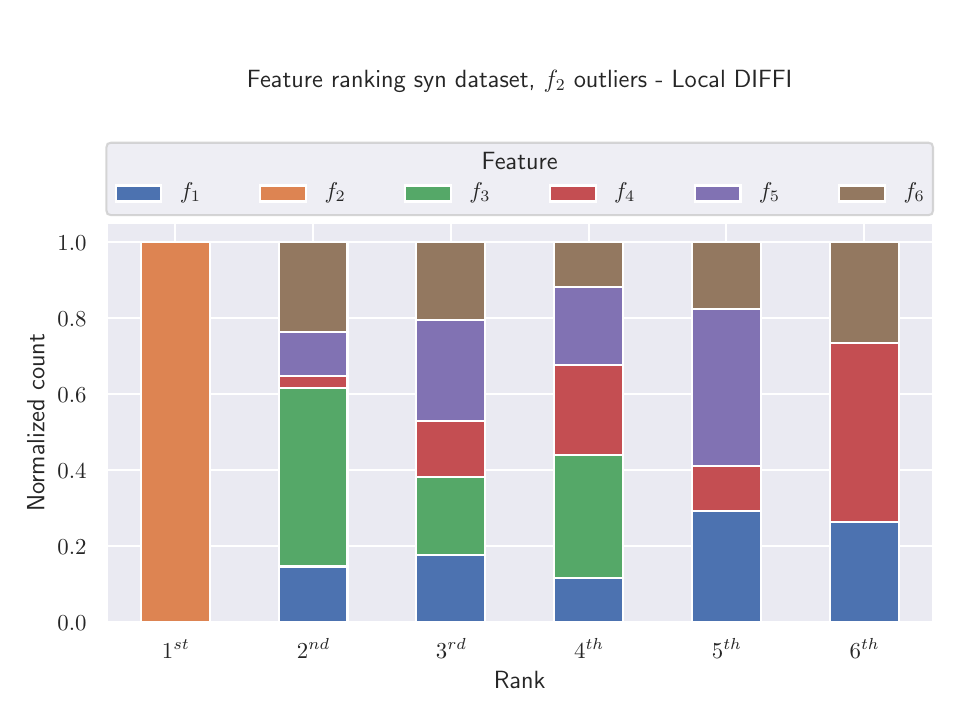}
	\end{subfigure}\hfill
	\begin{subfigure}[]{0.5\linewidth}
		\includegraphics[width=\textwidth]{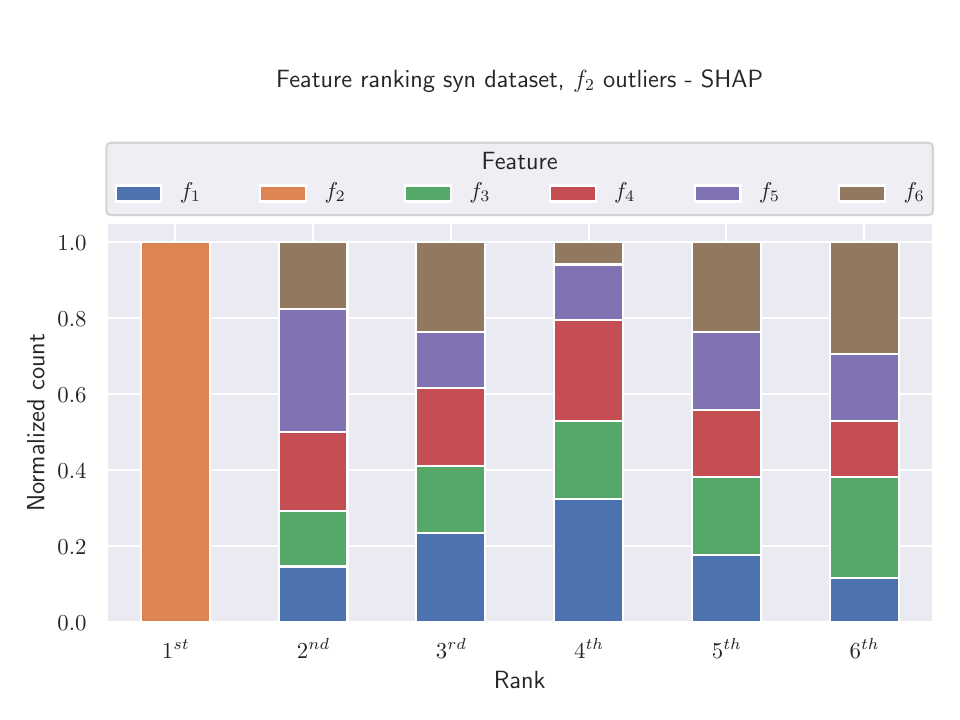}
	\end{subfigure}
	\begin{subfigure}[]{0.5\linewidth}
		\includegraphics[width=\textwidth]{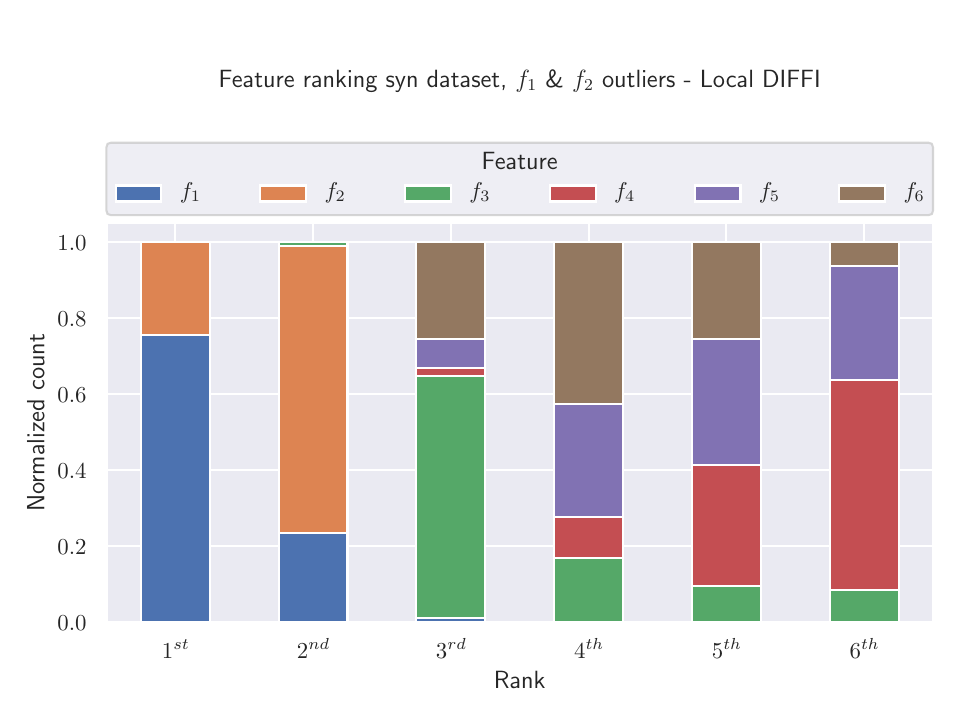}
	\end{subfigure}\hfill
	\begin{subfigure}[]{0.5\linewidth}
		\includegraphics[width=\textwidth]{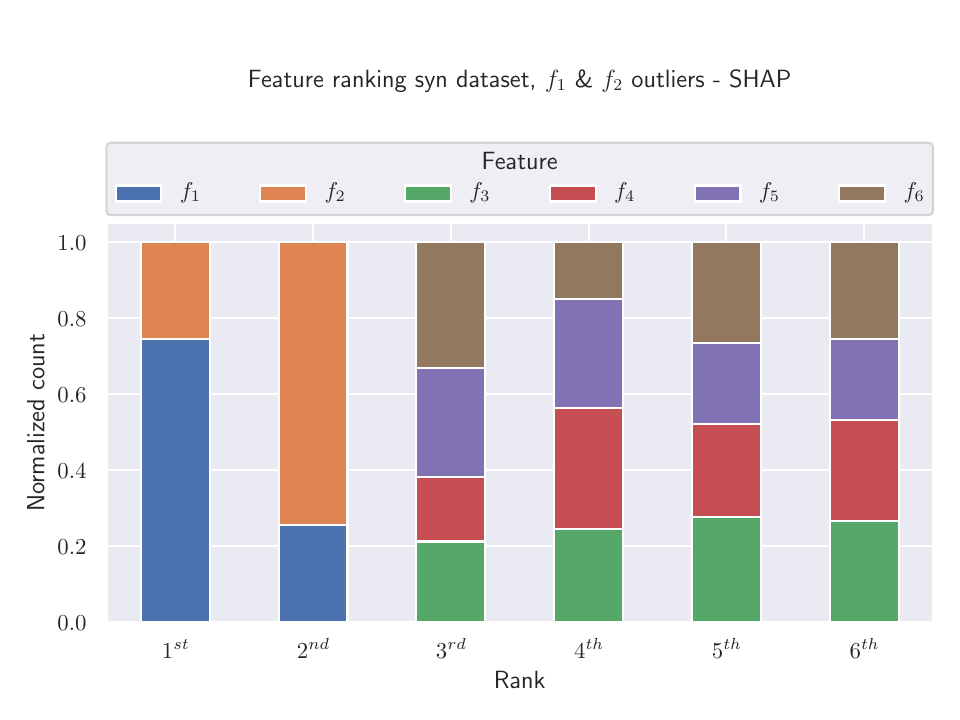}
	\end{subfigure}
	\caption{Feature rankings for the synthetic dataset based on local DIFFI scores (left column) and SHAP scores (right column): outliers on the $x$-axis (first row), on the $y$-axis (second row) and on the bisector (third row).}
	\label{fig:local_syn}
\end{figure}

Once obtained the predictions associated with the generated test outliers, we run the Local-DIFFI algorithm to get the corresponding local feature importance scores. We compared the performance of the Local-DIFFI method with SHAP. As can be seen in Figure \ref{fig:local_syn}, both methods perfectly identify the actual important feature(s): in the first two rows, for all correctly predicted outliers, the first column (representing the most important feature as estimated by the interpretability method) is always associated with the correct feature, namely $f_1$ and $f_2$ for outliers on the $x$-axis and on the $y$-axis, respectively; in the third row (referred to the points on the bisector), instead, both $f_1$ and $f_2$ are deemed important by Local-DIFFI and SHAP, thus aligning with prior knowledge. In this latter case, we also observed that feature importance scores provided by Local-DIFFI for feature $f_1$ and $f_2$ are comparable, while the same rightly does not happen for outliers on the axes. A major advantage of Local-DIFFI over SHAP is the computational time: while SHAP has an average execution time of 0.221 seconds per sample, Local-DIFFI runs in 0.023 seconds per sample on average. 


\subsubsection{Real-world dataset}

We consider a modified version of the Glass Identification UCI dataset\footnote{\url{https://archive.ics.uci.edu/ml/datasets/Glass+Identification}} originally intended for multiclass classification tasks. The dataset consists of 213 glass samples represented by a $9$-dimensional feature vector: one feature is the refractive index (RI), while the remaining features indicates the concentration of Magnesium (Mg), Silicon (Si), Calcium (Ca), Iron (Fe), Sodium (Na), Aluminum (Al), Potassium (K) and Barium (Ba). Originally the glass samples were representative of seven categories of glass type, but for our experiments we group classes 1, 2, 3 and 4 (i.e. window glass) to form the class of regular points, while the other three classes contribute to the set of anomalous data points (i.e. non-window glass): containers glass (class 5), tableware glass (class 6) and headlamps glass (class 7). We assess the performance of Local-DIFFI on predicted outliers belonging to class 7, considered as test data points. Similarly to \cite{gupta2018beyond}, we exploit prior knowledge on headlamps glass: the concentration of Aluminum, used as a reflective coating, and the concentration of Barium, which induces heat resistant properties, should be important features when distinguishing between headlamps glass and window glass.

\begin{figure}[tb]
\centering
\captionsetup[subfigure]{justification=centering}
	\begin{subfigure}[]{0.5\linewidth}
		\includegraphics[width=\textwidth]{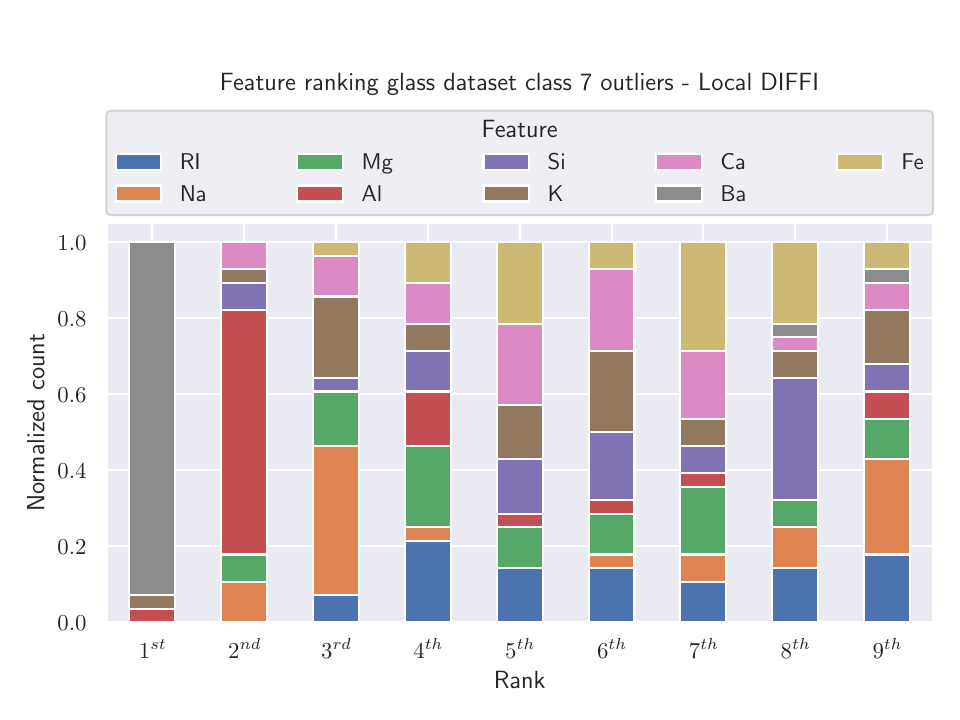}
	\end{subfigure}\hfill
	\begin{subfigure}[]{0.5\linewidth}
		\includegraphics[width=\textwidth]{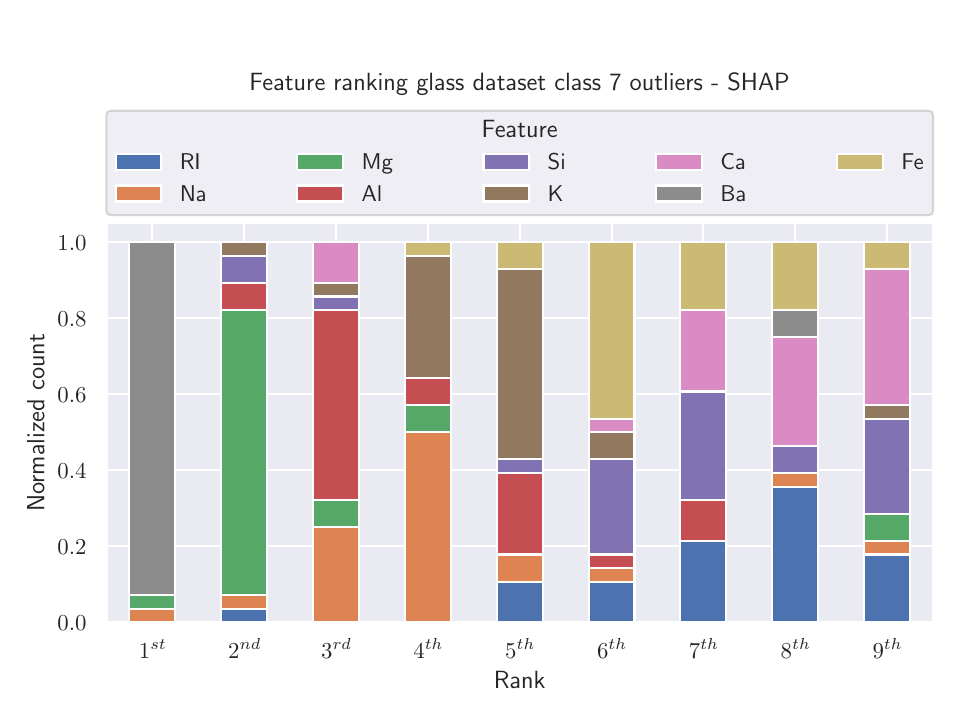}
	\end{subfigure}
	\caption{Feature rankings for the glass dataset based on local DIFFI scores (left column) and SHAP scores (right column): class 7 outliers (headlamps glass).}
	\label{fig:local_glass}
\end{figure}

We trained an instance of IF with 100 trees and $\psi = 64$, and obtained an F1-score on training data equal to 0.55. On the test data (class 7), the IF was able to identify 28 out of 29 anomalies. As for the synthetic dataset, we run the Local-DIFFI algorithm to get the local feature importance scores and compared the performance with those obtained with the SHAP method. As can be seen in Figure \ref{fig:local_glass}, Local-DIFFI identifies the concentration of Barium and Aluminum as the most important features in the vast majority of predicted anomalies, well aligned with a priori information about the task. The same cannot be said for SHAP: while the most important feature is still the concentration of Barium, the second most important feature for almost all predictions is the concentration of Magnesium. Additionally, also in this case the Local-DIFFI exhibits way smaller execution time (0.019 seconds per sample on average) than SHAP (0.109 seconds per sample on average).


\subsection{Unsupervised feature selection}

According to the procedure outlined in Section \ref{sec:DIFFI_fs}, we exploit the global DIFFI scores to define a ranking over the features representing the data points in the problem at hand. In all experiments described below we run $N_{fs}=5$ instances of IF.

To verify the quality of the selected features, we perform experiments on six common AD datasets from the Outlier Detection DataSets (ODDS) database\footnote{\url{http://odds.cs.stonybrook.edu/}}, whose characteristics are summarized in Table \ref{table:fs_datasets}. Once the ranking is obtained, we train an instance of IF by exploiting only the top $k$ most important features (according to the ranking), with $k=1, \dots, p-1$. We repeat the procedure 30 times (with different random seeds) and compute the median F1-score. 

\begin{table}[tb]
\centering 
\renewcommand*{\arraystretch}{1.2}
\caption{AD datasets used for unsupervised feature selection experiments.}\label{table:fs_datasets}
\footnotesize%
\begin{tabular}{l|cc}%
\toprule%
\textbf{Dataset ID} & \textbf{Num samples} & \textbf{Num features}\\%
\midrule%
\rowcolor{Gray}
\texttt{satellite} & 6435 & 36 \\%
\texttt{cardio} & 1831 & 21 \\\rowcolor{Gray}
\texttt{ionosphere} & 351 & 33 \\%
\texttt{lympho} & 148 & 18 \\\rowcolor{Gray}
\texttt{musk} & 3062 & 166 \\%
\texttt{letter} & 1600 & 32 \\\rowcolor{Gray}
\bottomrule%
\end{tabular}%
\end{table}

We provide comparisons with two other commonly used unsupervised feature selection techniques, i.e. Laplacian Score \cite{he2006laplacian} and SPEC \cite{zhao2007spectral}. We did not consider other techniques such as Nonnegative Discriminative Feature Selection \cite{li2012unsupervised} and $l_{2,1}$-Norm Regularized Discriminative Feature Selection \cite{yang2011l2} due to their prohibitive computational cost, which makes extensive usage practically cumbersome.

\begin{figure}[]
\centering
\captionsetup[subfigure]{justification=centering}
	\begin{subfigure}[]{0.5\linewidth}
		\includegraphics[width=\textwidth]{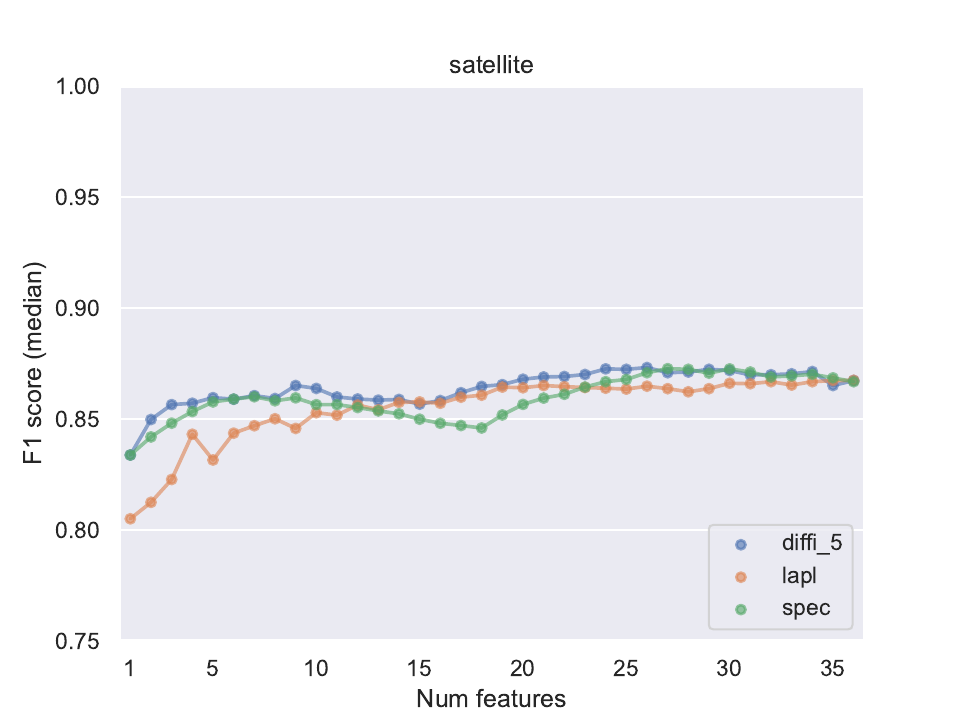}
	\end{subfigure}\hfill
	\begin{subfigure}[]{0.5\linewidth}
		\includegraphics[width=\textwidth]{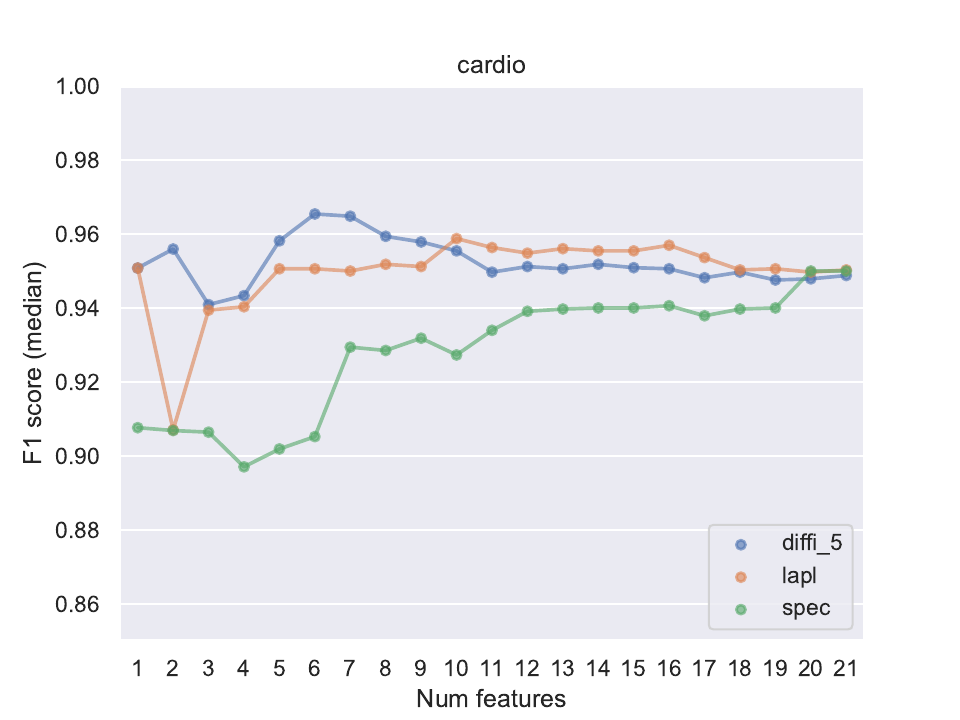}
	\end{subfigure}
	\begin{subfigure}[]{0.5\linewidth}
		\includegraphics[width=\textwidth]{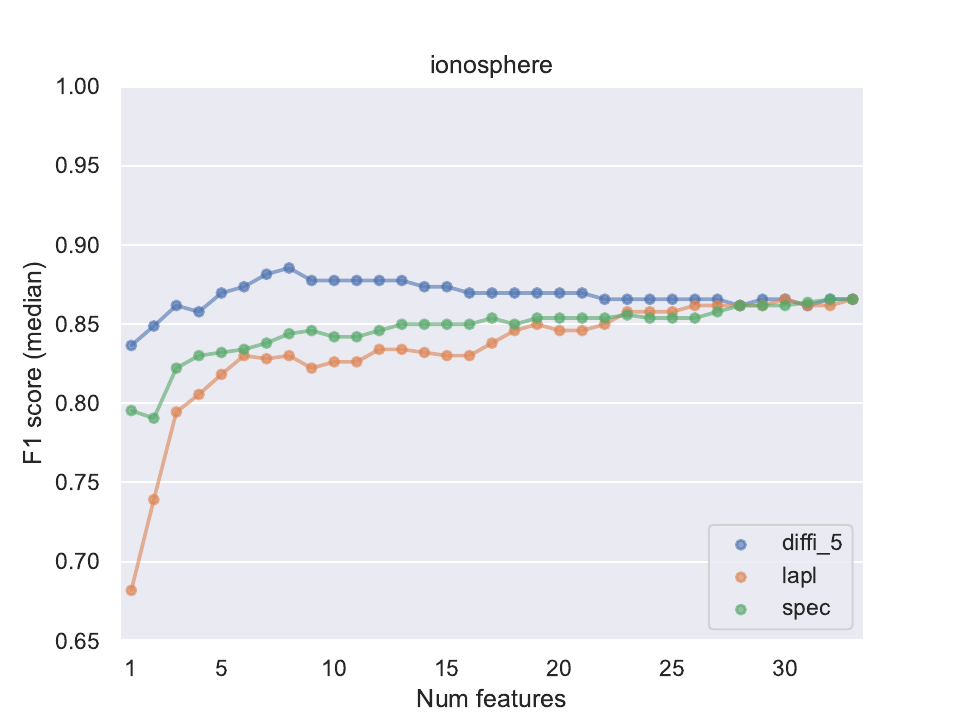}
	\end{subfigure}\hfill
	\begin{subfigure}[]{0.5\linewidth}
		\includegraphics[width=\textwidth]{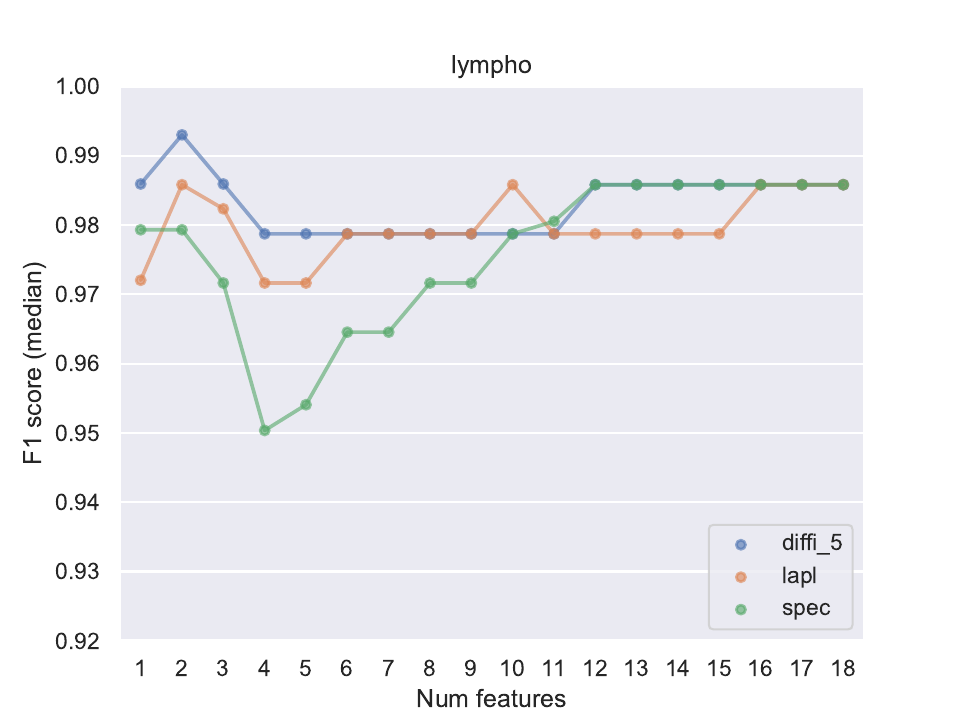}
	\end{subfigure}
	\begin{subfigure}[]{0.5\linewidth}
		\includegraphics[width=\textwidth]{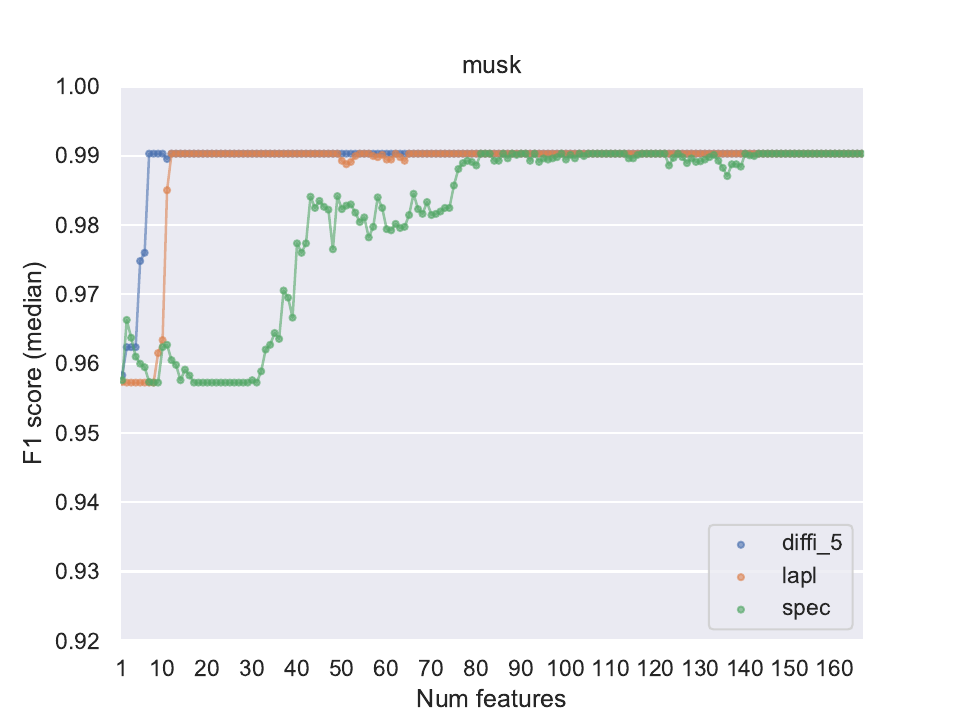}
	\end{subfigure}\hfill
	\begin{subfigure}[]{0.5\linewidth}
		\includegraphics[width=\textwidth]{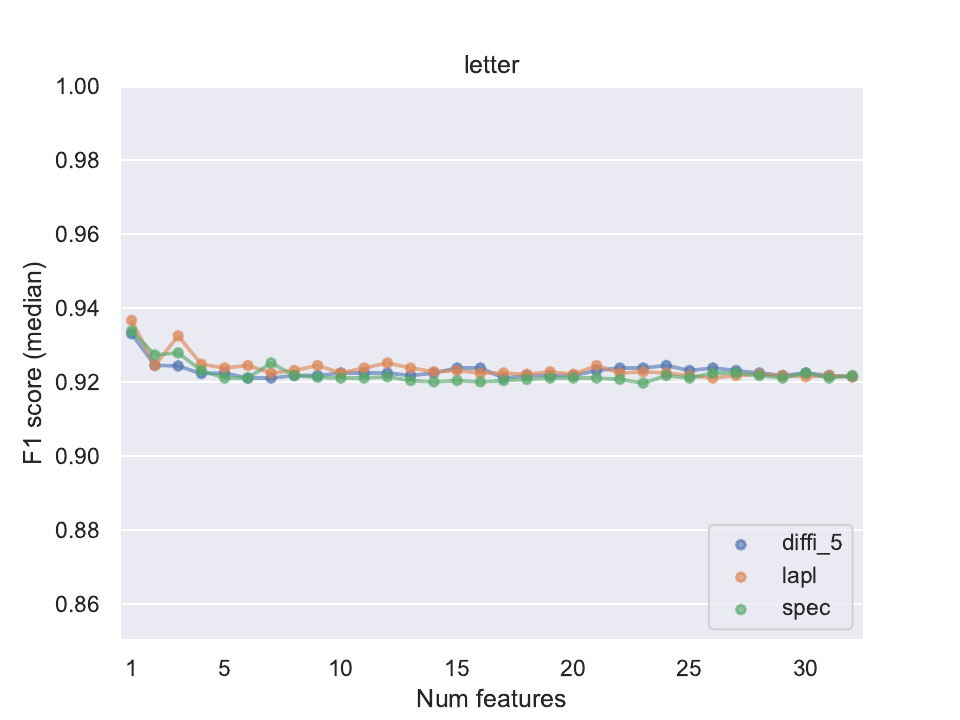}
	\end{subfigure}
	\caption{Evaluation of the global DIFFI method (\texttt{diffi\_5}) for unsupervised feature selection, compared with Laplacian Score (\texttt{lapl}) and SPEC (\texttt{spec}) methods.}
	\label{fig:fs}
\end{figure}

The hyperparameters values for the final IF model are tuned separately for each dataset through grid search, exploiting all the available features, and then kept fixed for all the experiments involving the same dataset. For the unsupervised feature selection methods, instead, we set the hyperparameters to the default values in order to be consistent with the goal of minimizing the user efforts in time-consuming operations. Furthermore, this approach perfectly fits real-world applications in which the lack of ground truth labels prevents the design of any principled hyperparameters tuning procedure, leading users to rely on existing heuristic rules.

As can be seen in Figure \ref{fig:fs}, the performance of DIFFI are comparable with those of the Laplacian Score and SPEC methods and consistently outperform them for a wide range of $k$ values (i.e. number of exploited features) for the \texttt{cardio}, \texttt{ionosphere}, and \texttt{musk} datasets. Also notice that in most cases DIFFI is able to identify the optimal combination of features, i.e. the subset of features leading to the highest (median) F1-score value.

Beyond these considerations, that are essentially of quantitative nature, it is equally - if not more - important to adopt the perspective of the final user and go through all the subtle aspects that make a specific methods more attractive than others. Along these lines, we believe that task-specific methods such as DIFFI are preferable over task-agnostic methods (like Laplacian Score and SPEC) as the features that actually relevant to solve a classification problem might not be relevant to solve an AD problem \cite{puggini2018enhanced}. This comes as no surprise in light of the different nature of the two tasks and it may unconsciously affect the user preference when reasoning about the most appropriate approach. Additionally, as mentioned in Section \ref{sec:DIFFI_fs}, the procedure based on DIFFI requires minimal - if any - hyperparameters tuning: the only hyperparameters are inherited from the underlying proxy model, i.e. an instance of IF, which has proved to provide satisfactory performance with the default hyperparameters values on a broad spectrum of applications.

\section{Conclusions and Future Works}
\label{sec:conclusion}

This paper introduces Depth-based Isolation Forest Feature Importance (DIFFI), a method to provide interpretabily traits to Isolation Forest (IF), one of the most popular and effective Anomaly Detection (AD) algorithm. By providing a quantitative measure of feature importance in the context of the AD task, DIFFI allows to describe the behavior of IF at global and local scale, providing insightful information that can be exploited by final users of an IF-based AD solution to get a better understanding of the underlying process and to enable root cause analysis. Moreover, the approach can help scientists and developers in improving their solutions by getting a better understanding of the important variables in their AD task. 

DIFFI is as effective as the current state-of-the-art method SHAP, with significantly smaller computational costs, making it really appealing for real world productive applications and even amenable for real-time scenarios. Moreover, we show that DIFFI can be employed to perform unsupervised feature selection, allowing the development of computationally parsimonious (and potentially more accurate) AD solutions. 

We believe that, given the exponentially growing interest in IF, DIFFI will be of paramount importance to enhance its usability and diffusion in real-world applications; we also believe that by equipping IF with DIFFI would lead to an increase in adoption in IF, thanks to the increased trust of users towards methods that have interpretability traits. 

Finally, we envision DIFFI to be possibly extended to other tree-based models for AD (for example the Extended Isolation Forest \cite{hariri2018extended}, the SCiForest \cite{liu2010detecting} or the Streaming HSTrees \cite{tan2011fast}). 
In particular, the low computational costs open up the opportunity to exploit DIFFI in the flourishing field of online AD applications with streaming data, where time efficiency is crucial~\cite{miao2018distributed, zhang2019online}.

\bibliographystyle{elsarticle-num-names} 
\bibliography{biblio}

\end{document}